\documentclass{article}
\usepackage{ijcai26}

\usepackage{times}
\usepackage{soul}
\usepackage{url}
\usepackage[hidelinks]{hyperref}
\usepackage[utf8]{inputenc}
\usepackage[small]{caption}
\usepackage{graphicx}
\usepackage{amsmath}
\usepackage{amsthm}
\usepackage{amssymb}
\usepackage{booktabs}
\usepackage{algorithm}
\usepackage{algorithmic}
\usepackage[switch]{lineno}
\usepackage{array}
\usepackage{tabularx}
\usepackage{adjustbox}
\usepackage{subcaption}
\usepackage{multirow}

\title{Optimizing Resource-Constrained Non-Pharmaceutical Interventions for Multi-Cluster Outbreak Control Using Hierarchical Reinforcement Learning}

\author{
Xueqiao Peng$^1$
\and
Andrew Perrault$^1$\\
\affiliations
$^1$The Ohio State University\\
\emails
peng.969@buckeyemail.osu.edu, perrault.17@osu.edu
}

\begin{document}

\maketitle

\begin{abstract}

Non-pharmaceutical interventions (NPIs), such as diagnostic testing and quarantine, are crucial for controlling infectious disease outbreaks but are often constrained by limited resources, particularly in the early outbreak stages.  In real-world public health settings, resources must be allocated across multiple outbreak clusters that emerge asynchronously, vary in size and risk, and compete for a shared resource budget. We define a cluster as a group of close contacts generated by a single infected index case. Thus, decisions must be made under uncertainty and heterogeneous demands while respecting operational constraints. We formulate this problem as a constrained restless multi-armed bandit and propose a hierarchical reinforcement learning framework. A global controller learns a continuous action cost multiplier that adjusts global resource demand, while a generalized local policy estimates the marginal value of allocating resources to individuals within each cluster. We evaluate the proposed framework in a realistic agent-based simulator of SARS-CoV-2 with dynamically arriving clusters. Across a wide range of system scales and testing budgets, our method consistently outperforms RMAB-inspired and heuristic baselines, improving outbreak control effectiveness by 20\%-30\%. Experiments on up to 40 concurrently active clusters further demonstrate that the hierarchical framework is highly scalable and enables faster decision-making than the RMAB-inspired method.

\end{abstract}

\section{Introduction}

The rapid emergence of novel infectious diseases has brought severe challenges to public health, particularly in the early stages of an outbreak when vaccines or targeted treatments are not available. In this stage, outbreak control mainly relies on non-pharmaceutical interventions (NPIs) such as diagnostic testing, contact tracing, and isolation. However, in practice, policymakers must make these decisions under considerable uncertainty. This uncertainty involves not only the potential state of the outbreak but also the policy objectives. For policy objectives, the relative importance of preventing transmission, avoiding unnecessary isolation, and reducing operational burden is often not clear beforehand and may change as the outbreak develops.

These difficulties are even more severe given the limited public health resources. For instance, when testing capacity is insufficient, policymakers must allocate testing based on conflicting needs without a clear standard. In reality, each confirmed case generates a local cluster of exposed individuals, and multiple clusters are often active simultaneously, competing for a shared testing budget. Heterogeneity in cluster size, timing, and symptom prevalence can lead to highly uneven and asynchronous testing demand both spatially and temporally~\cite{rahmandad2021behavioral,verelst2016behavioural}. Therefore, an effective early response to an outbreak requires a decision-making mechanism that can adapt to uncertainty while strictly adhering to resource constraints.

This setting naturally relates to Restless Multi-Armed Bandit (RMAB)~\cite{whittle1988restless}, but directly applying existing deep RMAB methods faces challenges. Most Deep RMAB solvers rely on network architectures (e.g., MLPs or RNNs) which require fixed-dimensional state inputs and a predefined static set of arms \cite{killian2021beyond}. This structural limitation makes them unable to handle the asynchronous nature of infectious diseases. Clusters dynamically emerge, resulting in the dimension of the global state space changing over time.  Moreover, classical and neural index-based methods often incur significant time-step computational overhead since the Whittle index must be computed or approximated iteratively at each time step \cite{weber1990index,nakhleh2021neurwin}, increasing computational cost per arm. In large-scale scenarios, this cost becomes prohibitive, making it difficult to perform rapid, low-latency decision-making to respond to an outbreak in real time.

In this work, we propose a hierarchical reinforcement learning framework to allocate the testing resources across multiple dynamically evolving clusters under a global budget. Instead of directly allocating tests, our method separates global coordination from local decision-making. A global controller adjusts the test usage across all the clusters through a continuous cost multiplier, which modulates the test cost perceived by the local policy. The local policy then evaluates the testing decisions at the individual level based on epidemiological features and cost signals provided by the global controller. The local clusters employ pre-trained Deep-Q Networks (DQNs) ~\cite{mnih2015human} enhanced with a Transformer~\cite{vaswani2017attention} architecture. We trained a generalized DQN that can produce different policies under different test cost signals. This design ensures local decisions remain effective under various objective trade-offs, allowing policymakers to adjust the policy according to their objectives. Final testing decisions are made by ranking candidate actions according to their estimated benefits and selecting the most valuable actions that fit within the available budget. This ensures that testing resources are consistently directed toward the most valuable decisions while remaining feasible under strict operational constraints.

We evaluate our framework using a realistic agent-based simulator of SARS-CoV-2 transmission. Across various testing budgets and outbreak scales, our method consistently outperforms the heuristic and RMAB-inspired baselines. Our main contributions are:
\begin{itemize}
\item We formulate the resource allocation problem as a constrained RMAB process with asynchronous cluster arrivals. This formulation bridges the gap between theoretical bandit models and real-world public health needs. In particular, it addresses the challenge of coordinating resources across temporally asynchronous outbreaks.

\item We propose a hierarchical reinforcement learning framework that decouples local intervention decisions from global resource allocation. A key contribution is the design of a generalized local DQN that can adapt to different resource constraints without retraining. By combining this with a global Proximal Policy Optimization (PPO) controller, our approach is able to dynamically adjust resource usage while maintaining computational efficiency.

\item Our proposed framework improves outbreak control effectiveness by 5-12\% relative to the RMAB-based methods and 20-30\% relative to heuristic strategies, while scaling to scenarios with up to $40$ concurrently active clusters and achieving approximately $5\times$ speedup in decision-making.
\end{itemize}

\section{Related Work}

\paragraph{Outbreak Control and NPIs.}
For a long time, research methods for outbreak control have included optimal control, compartment models, and large-scale simulations.\cite{nowzari2016analysis,brauer2017mathematical,probert2016decision}. While these classical methods offer important theoretical insights, they often rely on simplifying assumptions, such as population homogeneity or known dynamics. This limits their ability to support adaptive, refined decision-making under uncertainty. More recently, data-driven and reinforcement learning methods have been explored for epidemic control and mitigation policies~\cite{alamo2021data,kompella2020reinforcement}. Related bandit-based approaches have also studied public-health intervention and resource allocation problems~\cite{mate2020collapsing,killian2021beyond}. Building on this line of work, Peng et al.~\cite{peng2023using} studied individual-level testing and quarantine decisions under partial observability. However, most existing studies only consider a single population or a fixed decision-making context and do not explore how to coordinate limited resources to deal with multiple outbreak clusters that develop asynchronously over time.

\paragraph{Restless Multi-Armed Bandits (RMAB).}
Traditional RMAB solutions are typically based on index policies, the most well-known being the Whittle index\cite{whittle1988restless}. These methods require restrictive assumptions such as indexability and known transition dynamics \cite{glazebrook2006some,liu2010indexability,avrachenkov2022whittle}. RMAB formulations have been successfully applied to public health problems such as tuberculosis screening and maternal health outreach \cite{mate2020collapsing,biswas2021learn}. However, applying RMAB methods to outbreak response introduces additional challenges. Real-world outbreaks consist of clusters that appear asynchronously and vary widely in size, leading to state representations that have variable and time-varying dimensions, which are incompatible with standard deep RMAB architectures. Moreover, operational constraints such as daily testing capacity impose a tight per-timestep budget, whereas many existing approaches enforce constraints only in expectation.

Recent multi-agent reinforcement learning methods have addressed the scalability issue in large-scale populations \cite{yang2018mean,mao2026optimizing}. They often rely on decentralized execution and therefore cannot guarantee meeting global resource budgets. On the other hand, as the number of clusters increases, fully centralized methods face challenges related to the combinatorial action space and unstable training dynamics.

\paragraph{Hierarchical RL and Price-Based Coordination.}
Motivated by these challenges, our work draws on ideas from hierarchical reinforcement learning and price-based coordination, where the global signals regulate shared resources, while local policies focus on fine-grained decision-making \cite{barto2003recent,nachum2018data,vezhnevets2017feudal}. Related dual-based control mechanisms have been explored in networked and constrained systems, where Lagrange multipliers or shadow prices are used to manage global constraints \cite{bertsekas1997nonlinear,neely2010stochastic}. In contrast to prior approaches, we have explicitly designed a framework that can adapt to variable-sized, asynchronously arriving outbreak clusters and enforces strict budget constraints through a deterministic execution layer. By separating global resource regulation from local risk assessment, our approach achieves scalable coordination without incurring the combinatorial complexity of fully centralized control.

\section{Problem Description}
\label{sec: problem}
At the individual cluster level, intervention decisions such as testing and quarantine must balance epidemiological effectiveness against economic and social costs. In our setting, a cluster refers to a group of close contacts being exposed to a single index case simultaneously. The number of close contacts, demographic composition, and epidemiological characteristics of outbreaks may vary significantly. Following exposure, some contacts will become infected and develop symptoms, while others remain uninfected. Crucially, individuals’ true infection status is not directly observable; intervention decisions must be made sequentially under uncertainty.

Prior work by Peng~\shortcite{peng2023using} studies this \emph{single-cluster} optimization problem using reinforcement learning. Each cluster is modeled as an independent sequential decision process, with the objective defined as
\begin{equation}
\frac{-(S_1 + \alpha_2 S_2 + \alpha_3^{\text{true}} S_3)}{N},
\label{equ: reward}
\end{equation}
where $S_1$ denotes infectious days before quarantine, $S_2$ denotes unnecessary quarantine days for uninfected contacts weighted by $\alpha_2$, and $S_3$ denotes testing costs weighted by the true per-test cost $\alpha_3^{\text{true}}$. The objective is normalized by the cluster size $N$. Under the reward structure shown in Equation~\ref{equ: reward}, the quarantine policy has an optimal threshold form that is independent of the testing allocation. Therefore, we fix the quarantine policy and mainly focus on the resource allocation problem.

In this paper, we consider a more challenging scenario in which \emph{multiple} clusters evolve simultaneously and compete for a shared global testing budget. Clusters may activate and deactivate asynchronously over time, reflecting realistic outbreak dynamics across heterogeneous contact groups.

Let $\mathcal{A}_t$ denote the set of active clusters at time $t$. At each timestep, the global testing budget $B$ limits the total number of tests that can be performed across all active clusters. The system-level objective is to maximize the expected cumulative reward across clusters while respecting this per-timestep budget constraint:

\begin{equation}
\begin{aligned}
\max_{\pi} \quad
& \mathbb{E}\!\left[\sum_{t} \gamma^t 
\sum_{n \in \mathcal{A}_t} 
R_n\!\left(s_{n,t}, a_{n,t}\right)\right] \\
\text{s.t.}\quad
& \sum_{n \in \mathcal{A}_t} C_n(a_{n,t}) \le B, \quad \forall t .
\end{aligned}
\label{eq:global_problem}
\end{equation}
Here, $R_n$ denotes the cluster-level reward defined in Eq.~\eqref{equ: reward}, $C_n(a_{n,t})$ is the realized testing cost incurred by cluster $n$ at time $t$, and $B$ is the global testing budget.

This cross-cluster coupling introduces a system-level coordination challenge that cannot be solved by optimizing individual clusters alone. Directly learning the joint discrete resource allocation across clusters leads to an excessively large action space and poor scalability, while purely local strategies cannot take into account the global resource scarcity. Therefore, our goal is to design a scalable decision framework that can coordinate tests between heterogeneous, asynchronously activated clusters and strictly enforce the global test budget at each time step.

\section{Methods}
\subsection{RMAB Formulation and Lagrangian Relaxation}

The multi-cluster problem in Eq.~\eqref{eq:global_problem} can be formulated as a Restless Multi-Armed Bandit (RMAB)~\cite{whittle1988restless}, where each cluster is an arm that evolves independently and competes for a shared limited budget. Due to latent infection states, each cluster-level decision problem is partially observable and can be modeled as a Partially Observable Markov Decision Process (POMDP)~\cite{killian2021beyond}.

To handle global constraints on test resources, we adopt a Lagrangian relaxation approach, introducing a penalty parameter $\lambda$ to relax the budget constraint. For a fixed $\lambda$, the relaxed objective can be written as:

\begin{equation}
\small
\mathcal{L}(\lambda)
=
\mathbb{E}\!\left[
\sum_{t} \gamma^t
\sum_{n \in \mathcal{A}_t}
\big(R_n(s_{n,t},a_{n,t}) - \lambda C_n(a_{n,t})\big)
\right]
+ \frac{\lambda B}{1-\gamma}.
\end{equation}

This relaxation decomposes the global constrained problem into independent subproblems indexed by $\lambda$, each trading off reward against a penalized testing cost. Equivalently,
\begin{equation}
J(\lambda) = \frac{\lambda B}{1-\gamma} + \sum_{n} V_n(s_n,\lambda),
\end{equation}
where $V_n(s_n,\lambda)$ denotes the value of arm $n$ under the testing cost.

Rather than explicitly solving this dual optimization via linear programming, we use this formulation as a conceptual guide for our proposed framework.

Crucially, in our setting, the penalty parameter  $\lambda$ can be naturally interpreted as the cost coefficient $\alpha_3^{\text{active}}$ for each test, and is parameterized as:

\begin{equation}
\alpha_3^{\text{active}} = m_t \cdot \alpha_3^{\text{true}},
\end{equation}
where $m_t \ge 1$ is learned by a global controller.

This formulation implies that when the budget is tight, the controller increases $m_t$ to raise the effective test cost, thus suppressing testing actions in local policies.

\begin{figure*}[t]
\centering
\vspace{-2em}
\includegraphics[width=\textwidth]{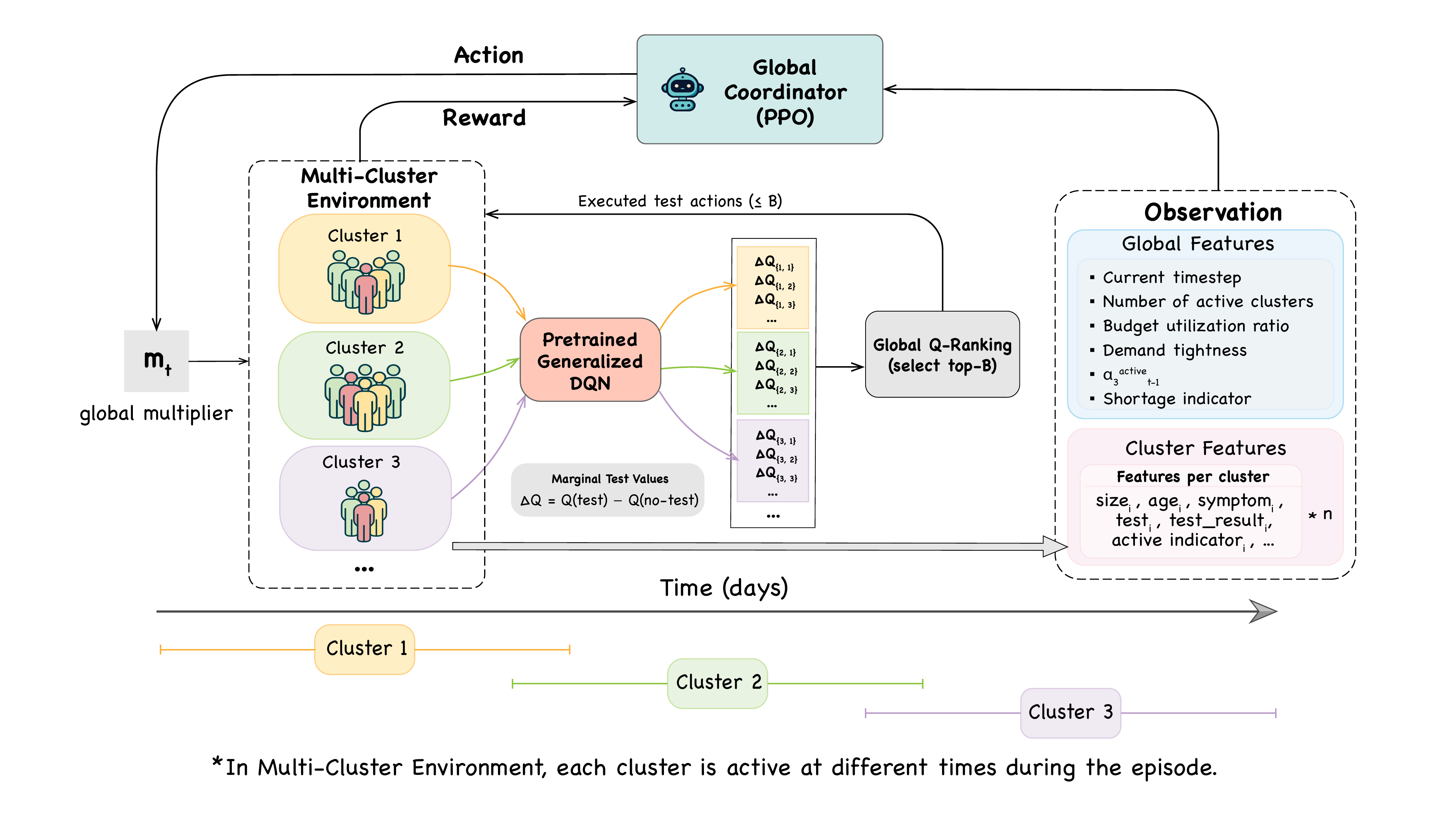}
\vspace{-2em}
\caption{Overview of the proposed hierarchical RL framework for multi-cluster outbreak control. A global PPO controller adjusts a shared testing cost multiplier, which modulates the perceived test cost used by a pretrained generalized DQN applied to each active cluster. Based on these cost-conditioned local value estimates, a global Q-ranking policy selects testing actions across clusters while strictly enforcing the hard global budget constraint.}
\vspace{-1.5em}
\label{fig: globalppo}
\end{figure*}

\subsection{Hierarchical Reinforcement Learning Framework}

To address the challenges of scalability and coordination in multi-cluster outbreak control, we propose a hierarchical reinforcement learning (HRL) framework. This framework separates \emph{global} allocation from \emph{local} intervention decision-making (As shown in Figure \ref{fig: globalppo}). Instead of directly allocating tests, the framework uses a global cost multiplier to adjust test costs, thereby adjusting resource usage, while local policies determine test priorities and make quarantine decisions within each cluster.

At each timestep, a pre-trained generalized local policy first evaluates marginal testing values for individuals in all active clusters based on close-contact features and the true testing cost, while quarantine decisions follow the threshold policy. These local evaluations generate a set of candidate testing actions. When the resulting testing demand exceeds the available budget, a global controller observes system-level statistics and outputs a scalar multiplier that scales the actual testing cost shared across clusters. This allows the controller to adjust resource usage without directly selecting individuals.

If the induced testing cost produced by the global controller still leads to testing demand that exceeds the budget, a deterministic global Q-ranking Policy is applied as an execution layer. Q-ranking aggregates candidate tests across clusters, ranks them by marginal value, and executes only the highest-value actions up to the budget limit. Even when changing the test cost is insufficient to meet the budget, this mechanism guarantees a strict budget while preserving local priorities. Together, these components achieve scalable coordination among heterogeneous clusters by decoupling global scarcity control from local decision-making and ensuring strict adherence to resource constraints at execution time.

\subsubsection{Local Policy: Generalized Transformer-Based DQN}

At the cluster level, we employ a generalized Deep Q-Network (DQN) \cite{mnih2015human} to estimate the marginal value of testing individual contacts. To handle the dynamic number of individuals within each cluster, we employ a Transformer~\cite{vaswani2017attention} in DQN. Quarantine decisions follow a threshold policy adopted from  work~\cite{peng2023using}, and the learnable decision is whether to administer a diagnostic test  (\texttt{test} vs.\ \texttt{no-test}) to each individual.

A key challenge is that optimal testing depends on the actual test cost, which will affect the test usage in each cluster. Rather than retrain separate policies for different cost regimes, we use a \emph{generalized} DQN that conditions on the perceived test cost.

Formally, the observation for cluster $n$ at time $t$ is:
\[
o_{n,i,t} = [\, s_{n,i,t},\; \alpha^{\text{active}}_{3,t}],
\]
where $s_{n,i,t}$ summarizes belief-based features (e.g., infection probabilities, recent symptoms, and testing history). The full details are provided in Appendix ~\ref{app:local_dqn}.

The DQN outputs action-values:
\[
Q_{\phi}(o_{n,i,t}, a), \quad a \in \{\texttt{no-test}, \texttt{test}\}.
\]

To ensure consistent behavior of the learned policy, we introduce a gradient-based regularization term. From a public health decision-support perspective, this enforces a basic economic and clinical principle: when the actual cost of testing increases, the tendency to recommend testing should not increase. Without such a constraint, deep reinforcement learning agents may exhibit counterintuitive behaviors due to function approximation errors, such as performing more tests under higher costs. By explicitly enforcing monotonicity with respect to the testing cost, we improve the logical consistency of the learned policy. This is widely considered an important requirement for interpretable and trustworthy AI systems in the healthcare field \cite{holzinger2019causability}.

The total loss at each training step is defined as:
\begin{equation}
\mathcal{L}_{\text{total}} = \mathcal{L}_{\text{TD}} + \lambda_{\text{gp}} \cdot \mathcal{L}_{\text{grad}},
\end{equation}
where $\mathcal{L}_{\text{TD}}$ is the standard temporal-difference loss,
and $\mathcal{L}_{\text{grad}}$ is the gradient penalty term:
\begin{equation}
\mathcal{L}_{\text{grad}}
= \mathbb{E}_{o_t,\,a_t}\!\left[
      \Bigl(
        \max\!\Bigl(0,\;\frac{\partial Q(o_t,a_t)}{\partial \alpha_3}
                       \;-\; g_{\text{target}}\Bigr)
      \Bigr)^{\!2}
    \right]
\end{equation}
Here, $g_{\text{target}} < 0$ is a target gradient that encourages decreasing Q-values with higher $\alpha_3$. The coefficient $\lambda_{\text{gp}}$ controls the strength of this regularization.

This regularization enables the local generalized DQN not only to learn the dynamics of the outbreak but also to make a smooth and predictable response to changes in the scarcity of detection resources. After training, the policy adapts continuously to different $\alpha_3$ inputs: higher values suppress testing, while lower values encourage more aggressive testing behavior. We emphasize that the environment reward is always computed using the true per-test cost $\alpha^{\text{true}}_3$, while the local DQN takes $\alpha^{\text{active}}_{3,t}$ as input. The $\alpha^{\text{active}}_{3,t}$ is provided by the global controller to control the test usage within each cluster.

\subsubsection{Global Policy: PPO-Based Controller}
To coordinate testing across dynamically activated clusters, we introduce a global controller that adjusts the test usage across the system using time-varying cost signals.
The controller is implemented using Proximal Policy Optimization (PPO)~\cite{schulman2017proximal} and operates at the system level, rather than selecting test actions for individual contacts.

The PPO strategy outputs a scalar multiplier $m_t$, which parameterizes the active test cost coefficient defined in the RMAB formula. This multiplier allows the controller to adjust the overall detection demand from each local cluster, while remaining independent of individual-level decisions. The environment evaluates rewards using the fixed true per-test cost $\alpha_3^{\text{true}}$.

Importantly, the controller does not directly enforce feasibility.
When total testing demand is below the available budget, the final allocation is identical to that under the true cost, regardless of the specific value of $m_t$.
When demand would exceed capacity, increasing $m_t$ will increase perceived test costs and suppress demand, while the subsequent ranking-based allocation mechanism ensures strict adherence to hard budget constraints.

At each time step $t$, the PPO policy operates based on a compressed observation, which includes system-level information and cluster features. This enables scalable decision-making as clusters emerge. Full details of the controller observation space and feature construction are provided in Appendix ~\ref{app:global_obs}.

\subsubsection{Global Q-Ranking for Test Execution}

At each decision step, the generalized local DQN evaluates testing decisions for all individuals in all active clusters under the current perceived test cost. For each individual, the DQN outputs action-value estimates for two actions: \texttt{test} and \texttt{no-test}.
The difference between these values,
\[
\Delta Q_{n,i} = Q_{\phi}(o_{n,i,t}, \texttt{test}) - Q_{\phi}(o_{n,i,t}, \texttt{no-test}),
\]
represents the marginal benefit of administering a test, with larger values indicating higher testing priority.

Because all local Q-values are conditioned on the same perceived test cost, these marginal scores are directly comparable across clusters. To enforce the hard global testing budget, we apply a global Q-ranking policy: all candidate testing actions from active clusters are pooled and ranked by $\Delta Q$ in descending order. Tests are executed for the top-ranked candidates with positive marginal value, up to the available budget $B$ at the current timestep. If fewer than $B$ candidates have positive marginal value, the remaining budget is left unused, and all other individuals are assigned \texttt{no-test}. This procedure guarantees feasibility by construction; full details are provided in Appendix~\ref{app:qranking}.

Global Q-ranking serves as a deterministic execution layer that separates marginal value estimation from budget enforcement. This decomposition avoids the instability of directly learning discrete, globally constrained allocations while remaining interpretable, as tests are allocated to individuals with the highest estimated marginal benefit under the current cost regime.

\section{Experiments}
\subsection{Experimental Setup}

We evaluate the proposed framework using a realistic agent-based simulator of SARS-CoV-2 transmission that models infection, testing, and isolation at the individual level, allowing intervention decisions to directly affect disease dynamics and future observability. Epidemiological parameters are drawn from public health literature, and the simulator explicitly captures high-transmission scenarios through a small fraction of highly infectious index cases, reflecting the heavy-tailed risk observed in real outbreaks.

The environment is organized as a multi-cluster system induced by contact tracing, where each confirmed case generates a cluster of exposed individuals evolving independently over time. Multiple clusters may be active concurrently and compete for a shared global testing budget, while all transmission and intervention dynamics remain individual-level.

At each timestep, a hard global testing budget constrains total test execution across clusters. During training, the budget is randomized to improve robustness across resource regimes. We evaluate performance under both synchronous and asynchronous cluster activation using the cumulative multi-objective reward in Equation~\ref{equ: reward}, normalized by cluster size. Results are averaged over five random seeds, with full experimental details provided in Appendix~\ref{app:exp_details}. Code and Environment are published at ~\url{https://anonymous.4open.science/r/ttesfjwjm-16}.

\subsection{Comparison Policies}

We compare our hierarchical reinforcement learning approach against several baseline policies for allocating limited testing resources under a hard global budget constraint. All methods are evaluated in the same environment and strictly enforce feasibility at every timestep.

Across all methods, testing and quarantine decisions are decoupled. Except for Symp-AvgRand, all baselines apply the same fixed threshold-based quarantine policy used throughout this paper (Appendix~\ref{app:local_dqn}); Symp-AvgRand instead uses symptom-based quarantine as a representative public health heuristic. Unless otherwise stated, all experiments use a fixed quarantine cost coefficient $\alpha_2 = 0.1$ and a true per-test cost $\alpha_3^{\text{true}} = 0.05$.

\textbf{Symp-AvgRand.} The global testing budget is evenly divided across active clusters, and individuals are selected uniformly at random for testing. Quarantine is based on observed symptoms and positive test results.

\textbf{Thres-AvgRand.} Testing resources are evenly allocated across clusters, with individuals selected uniformly at random for testing. Quarantine follows the threshold-based policy used throughout this paper.

\textbf{Thres-SizeRand.} Testing resources are allocated proportionally to cluster sizes, while individuals within each cluster are selected uniformly at random for testing. Quarantine again follows the threshold policy.

\textbf{Fixed-$M$-QR.} A fixed global cost multiplier ($M{=}1$) is applied to the pretrained generalized DQN, inducing a constant perceived testing cost across all timesteps. This baseline evaluates whether a static scarcity signal is sufficient for coordinating testing decisions. Results with alternative fixed $M$ values are reported in Appendix~\ref{app: fixed_m_ablation}.

\textbf{Bin-$M$-QR.} The global cost multiplier is adjusted online via binary search to satisfy the testing budget at each timestep. This method provides a reactive, non-learning coordination mechanism, analogous in spirit to Whittle-index–style approaches, but without requiring indexability or explicit index computation.

\textbf{Hierarchical PPO (Ours).} Our approach learns a time-varying global cost multiplier using a PPO-based controller, enabling adaptive coordination of testing decisions based on the evolving global system state.

All learning-based baselines share the same pretrained generalized local DQN and differ only in how the cost multiplier is selected. Full implementation details for all comparison policies are provided in Appendix~\ref{app: policies}.

\subsection{Generalized vs.\ Fixed DQN}
Table ~\ref{tab:gen_vs_fixed_dqn} shows the results comparing the generalized DQN with fixed-cost DQN policies. The generalized DQN is trained with $\alpha_3$ randomly sampled from $[0.0, 0.1]$, while each fixed-cost DQN is trained under a single fixed $\alpha_3$ value. Both generalized and fixed-cost DQN models use a fixed quarantine cost $\alpha_2$ = 0.1.

Across all evaluated settings, the generalized policy achieves performance comparable to that of specialized fixed-cost policies. This confirms that a single conditional model can effectively capture the optimal policy across varying cost regimes without sacrificing precision. This result has significant practical implications. Rather than training and maintaining a suite of separate models for every possible cost scenario, a single generalized model can be deployed to handle a wide range of resource conditions.

This substantially reduces computational overhead, as policy adaptation to different resource conditions requires only changing the input cost parameter rather than retraining the model. Moreover, the generalized formulation enables efficient what-if analysis. Policymakers can rapidly evaluate how testing behavior and outcomes would change under alternative cost or budget assumptions using a single trained model. This flexibility is particularly valuable in real-world public health settings, where resource constraints can shift over time and rapid scenario evaluation is often required.

\begin{table}[htbp]
  \centering
  \small
  \setlength{\tabcolsep}{1pt}      
  \renewcommand{\arraystretch}{1}
  \begin{tabularx}{\columnwidth}{l *{4}{>{\centering\arraybackslash}X}}
    \toprule
        & $\alpha_3{=}0.01$ & $\alpha_3{=}0.05$ & $\alpha_3{=}0.08$ & $\alpha_3{=}0.10$\\
    \midrule
    Fixed-DQN & $-0.23{\pm}0.02$ & $-0.33{\pm}0.02$ & $-0.42{\pm}0.02$ & $-0.50{\pm}0.04$ \\
    Gen-DQN & $-0.25{\pm}0.02$ & $-0.34{\pm}0.02$ & $-0.41{\pm}0.02$ & $-0.47{\pm}0.03$ \\
    \bottomrule
  \end{tabularx}
  \vspace{-0.5em}
   \caption{Generalized vs. Fixed DQN performance under varying test penalty cost ($\alpha_3$). Results indicate that the generalized policy achieves comparable performance, demonstrating adaptability to different test penalties.}
   \vspace{-1.5em}
   \label{tab:gen_vs_fixed_dqn}
\end{table}

\subsection{Analysis}

\begin{table*}[t]
\centering

\begin{adjustbox}{width=\textwidth}
\begin{tabular}{lcccccc}
\toprule
& Symp-AvgRand & Thres-AvgRand & Thres-SizeRand & Fixed-$M$-QR(1) &Bin-$M$-QR   & Hier-PPO \\
\midrule
\#C=10, \#B=20 & $-2.12{\pm}0.03$ & $-1.02{\pm}0.01$ & $-0.93{\pm}0.02$ & $-0.64{\pm}0.01$ & $-0.61{\pm}0.01$ & $\mathbf{-0.56{\pm}0.01}$ \\
\#C=10, \#B=200 & $-2.64{\pm}0.03$ & $-2.04{\pm}0.01$ & $-2.12{\pm}0.01$ & $-0.67{\pm}0.01$ & $-0.66{\pm}0.02$ & $\mathbf{-0.63{\pm}0.01}$  \\
\#C=20, \#B=40 & $-2.31{\pm}0.04$ & $-1.11{\pm}0.01$ & $-1.01{\pm}0.02$ & $-0.65{\pm}0.01$ & $-0.64{\pm}0.01$ & $\mathbf{-0.57{\pm}0.01}$ \\
\#C=20, \#B=400 & $-2.79{\pm}0.02$ & $-2.15{\pm}0.01$ & $-2.21{\pm}0.01$ & $-0.68{\pm}0.01$ & $-0.68{\pm}0.01$ & $\mathbf{-0.63{\pm}0.01}$  \\
\#C=40, \#B=80 & $-2.34{\pm}0.03$ & $-1.13{\pm}0.01$ & $-1.02{\pm}0.01$ & $-0.66{\pm}0.01$ & $-0.63{\pm}0.02$ & $\mathbf{-0.59{\pm}0.01}$ \\
\#C=40, \#B=800 & $-2.82{\pm}0.02$ & $-2.18{\pm}0.06$ & $-1.57{\pm}0.02$ & $-0.78{\pm}0.01$ & $-0.68{\pm}0.02$ & $\mathbf{-0.64{\pm}0.01}$  \\
\bottomrule
\end{tabular}
\end{adjustbox}

\vspace{-0.5em}
\caption{Performance comparison of various test allocation methods under asynchronous cluster activation environments. Results highlight the advantage of the proposed hierarchical allocation PPO (Hier-PPO) method, particularly in asynchronous settings where resource demand varies over time. In synchronous settings, value-based policies exhibit similar performance. (\#C: number of clusters, \#B: daily testing budget)}
\vspace{-1em}
\label{tab:overall_result}
\end{table*}

Table~\ref{tab:overall_result} summarizes performance under asynchronous cluster activation across different cluster numbers and testing budgets. Hier-PPO achieves the highest average return across all settings. Compared with the strongest reactive baseline, Bin-$M$-QR, Hier-PPO improves overall returns by approximately $5\%$–$12\%$ under low-to-moderate budgets. Moreover, all globally coordinated Q-ranking methods (Fixed-$M$-QR, Bin-$M$-QR, and Hier-PPO) substantially outperform heuristic baselines, yielding $20\%$–$30\%$ improvements over threshold-based random allocation in representative constrained settings (e.g., around $-1.11$ vs. $-0.65$ for $\#C$=20,$\#B$=40). When budgets are loose relative to demand, performance gaps among globally coordinated methods narrow, consistent with diminishing returns from scarcity control.

To understand why these results arise, Table~\ref{tab:async_s123} decomposes performance into $S_1$ (unquarantined infections), $S_2$  (unnecessary quarantine), and $S_3$ (testing cost). Under tight budgets ($\#B$=40), Symp-AvgRand exhibits very high $S_1$, indicating that quarantine based on symptom and test results will miss many asymptomatic infections. Thres-AvgRand and Thres-SizeRand reduce $S_1$ but incur substantially larger $S_3$ and higher $S_2$, showing that they rely on heavy testing and quarantine lots of uninfected individuals. In contrast, Q-ranking methods keep $S_3$ close to the effective budget limit (around $3$–$4$) while achieving much lower $S_1$, demonstrating that global prioritization can reduce unquarantined infections without excessive testing. Among these, Hier-PPO attains the lowest $S_1$ and $S_2$, indicating more accurate identification of high-impact tests, thus performing more accurate quarantine.

Under loose budgets ($\#B$=400), threshold-based random allocation achieves very low $S_1$ by testing nearly everyone, but this comes with extremely high testing cost ($S_3 \approx 36$–$38$). In contrast, Q-ranking methods refrain from exhausting available capacity. Because there is a cost for testing,  they perform tests only when the estimated marginal benefit is positive, resulting in substantially lower $S_3$ (around $4$) with only a modest increase in $S_1$. This reflects a deliberate cost–benefit trade-off rather than suboptimal behavior.

Table~\ref{tab:async_s123} further highlights the distinction between Bin-$M$-QR and Hier-PPO. Both methods intervene only when the proposed testing demand exceeds the per-timestep budget. However, Bin-$M$-QR relies on binary search to adjust the cost multiplier reactively at each timestep, whereas Hier-PPO learns a smooth, state-dependent adjustment policy. This learned policy better aligns the scarcity signal with the evolving system state when constraints bind, yielding consistently lower $S_1$ and $S_2$ at comparable testing cost.

Overall, these results demonstrate that global prioritization is critical for multi-cluster outbreak control and that learning how to adjust cost signals under binding constraints leads to more efficient and stable allocations than purely reactive strategies.

\begin{table}[htbp]
  \centering
  \footnotesize
  \setlength{\tabcolsep}{1pt}
  \renewcommand{\arraystretch}{1.1}
  \begin{tabularx}{\columnwidth}{l c *{3}{>{\centering\arraybackslash}X}}
    \toprule
    Method & $(\#C,\#B)$ & S1 & S2 & S3 \\
    \midrule
    Symp-AvgRand      & \multirow{6}{*}{$(20,40)$}  & $1.64{\pm}0.04$  & $0.44{\pm}0.02$  &  $12.54{\pm}0.09$ \\
    Thres-AvgRand    &                             & $0.33{\pm}0.01$  & $1.59{\pm}0.03$  &  $12.54{\pm}0.09$ \\
    Thres-SizeRand &                             & $0.39{\pm}0.01$  & $1.58{\pm}0.03$  &  $9.33{\pm}0.06$ \\
    Fixed-$M$-QR     &                             & $0.35{\pm}0.01$  & $1.02{\pm}0.02$  &  $3.45{\pm}0.05$ \\
    Bin-$M$-QR        &                             & $0.36{\pm}0.01$  & $0.99{\pm}0.02$  &  $3.35{\pm}0.04$ \\
    Hier-PPO       &                             & $0.30{\pm}0.04$  & $0.95{\pm}0.02$  &  $3.41{\pm}0.06$ \\
    \midrule
    Symp-AvgRand     & \multirow{6}{*}{$(20,400)$} & $0.94{\pm}0.02$ & $0.58{\pm}0.01$ & $35.95{\pm}0.01$ \\
    Thres-AvgRand     &                             & $0.17{\pm}0.01$ & $1.81{\pm}0.04$ & $35.95{\pm}0.05$ \\
    Thres-SizeRand &                             & $0.15{\pm}0.01$ & $1.76{\pm}0.04$ & $37.71{\pm}0.05$ \\
    Fixed-$M$-QR      &                             & $0.37{\pm}0.01$ & $1.09{\pm}0.03$ & $4.07{\pm}0.04$ \\
    Bin-$M$-QR        &                             & $0.36{\pm}0.01$ & $1.08{\pm}0.03$ & $4.08{\pm}0.04$ \\
    Hier-PPO       &                             & $0.30{\pm}0.01$ & $0.99{\pm}0.02$ & $4.10{\pm}0.03$ \\
    \bottomrule
  \end{tabularx}
  \vspace{-0.5em}
  \caption{Comparison of different methods under asynchronous settings for $\#C{=}20$ with two global budgets. Results are reported for metrics $S_1$, $S_2$, and $S_3$.}
  \vspace{-1.4em}
  \label{tab:async_s123}
\end{table}

\paragraph{Computational efficiency.}
In addition to improved performance, Hier-PPO is computationally more efficient at decision time, reducing runtime by approximately $4-8\times$ compared with Bin-$M$-QR. Hier-PPO directly outputs the test cost multiplier through a single forward pass of the global PPO, whereas Bin-$M$-QR determines the multiplier via iterative binary search and therefore requires multiple demand evaluations per timestep. A detailed analysis of per-timestep decision-time runtime across different numbers of active clusters is provided in Appendix~\ref{app:runtime}.

\section{Discussion}

This work investigates how limited testing resources can be allocated across multiple clusters under per-timestep budget constraints. By decoupling system-level coordination from local decision-making, the proposed hierarchical framework provides a scalable and practical approach for multi-cluster outbreak control that aligns with constraints faced in real-world public health settings.

The key advantage of this framework is its separation of responsibilities across decision layers. The global controller regulates the overall testing demand through an interpretable cost signal, while the local policy focuses on estimating the marginal value of testing at the individual level. This design avoids the combinatorial complexity of directly optimizing global allocations and enables stable behavior across different cluster sizes, activation patterns, and budget regimes. In practice, this design leads to favorable trade-offs in which missed infections and unnecessary quarantines are reduced simultaneously, because testing resources are used on the most valuable individuals.

This framework uses a threshold-based quarantine policy rather than jointly learning testing and quarantine decisions. Under the POMDP formulation and reward structure considered in this paper, the threshold policy is provably optimal ~\cite{peng2023using}. Fixing the quarantine policy provides theoretical guarantees and improves training stability in large, multi-cluster environments. Furthermore, the quarantine cost parameter $\alpha_2$ also plays an important role. By determining the relative penalty of unnecessary quarantine, $\alpha_2$ shapes the value of information provided by testing. When $\alpha_2$ is very small, quarantine is inexpensive and applied to every contact, thus limiting the value of testing since everyone is quarantined regardless of test results. On the other hand, when $\alpha_2$ is large, quarantine becomes costly and rarely considered, again making testing less informative since testing results do not trigger any meaningful action. For this reason, the main experiments focus on the setting with $\alpha_2 = 0.1$. We also extended the model to a setting where both quarantine and testing costs are generalized, as discussed in Appendix ~\ref{app:gen_both}.

Several limitations point to important directions for future work. First, although the global cost signal provides an interpretable summary of system-level scarcity, local testing decisions are produced by neural networks and are not directly interpretable at the individual level. This lack of transparency is a common challenge for reinforcement learning methods applied to public health decision support and may limit trust and adoption in practice \cite{amann2020explainability}. Improving interpretability and alignment with public health reasoning remain important directions.

Second, the framework omits behavioral factors such as imperfect compliance with testing and quarantine guidance. In real outbreaks, adherence varies over time due to fatigue, risk perception, and trust in public health messaging, which can substantially affect intervention outcomes \cite{funk2010modelling,verelst2016behavioural}. Modeling such behavioral dynamics remains an open challenge. Recent advances in large language models offer a promising direction for incorporating realistic patterns of human decision-making and compliance inferred from unstructured data \cite{park2023generative,argyle2023out}.

Overall, this work demonstrates the potential of hierarchical reinforcement learning as a decision-support tool for outbreak response under resource constraints. Rather than replacing human judgment, the proposed framework is designed to support public health practitioners in reasoning about trade-offs, prioritizing testing, and adapting interventions. We view this approach as a step toward AI systems that enable more informed, timely, and accountable decision-making in real-world public health settings.

\bibliographystyle{named}
\bibliography{ijcai26}

\clearpage
\appendix
\setcounter{table}{0}
\renewcommand{\thetable}{A.\arabic{table}}

\setcounter{figure}{0}
\renewcommand{\thefigure}{A.\arabic{figure}}

\section{Model Details}
\label{app:model_detail}

\subsection{Local Generalized DQN}
\label{app:local_dqn}

At the cluster level, we adopt the single-cluster reinforcement learning framework introduced by Peng~\shortcite{peng2023using} as the local decision module.
In this framework, each cluster is controlled by a Deep Q-Network (DQN) that makes individual-level testing decisions under partial observability.
Our implementation extends this framework to support adaptation across different testing cost regimes without retraining.

\paragraph{Supervised Learning Encoder}
Because true infection states are unobserved at decision time, the local DQN relies on a supervised learning (SL) encoder to construct belief-based features. The SL encoder maps partial and noisy observations into an estimate of individual infection risk, enabling the downstream policy to reason about heterogeneous and potentially highly transmissive clusters.

For each individual, the SL encoder takes as input observable information that would be available to a public health decision-maker, including symptom indicators and recent symptom history, testing history and test outcomes (when available), and time-since-exposure or time-since-cluster-activation features. The encoder estimates the probability that the individual is currently infected. This probability is used as an input feature to the local DQN and also supports threshold-based quarantine decisions.

\paragraph{Threshold-based quarantine policy.}
Across all methods, including learned baselines, quarantine decisions follow a threshold-based policy adopted from Peng~\shortcite{peng2023using}. Specifically, an individual is quarantined whenever the predicted infection probability produced by the SL encoder exceeds a predefined threshold. 

Let $\mathcal{O}^{(t)}$ denote all symptom observations available within a cluster up to day $t$. Based on these observations, the belief that individual $n$ is infected at time $t$ is given by
\[
q_n^{(t)} := \mathbb{P}\!\left(i_n^{(t)} = 1 \mid \mathcal{O}^{(t)}\right).
\]

Under the assumed reward structure, quarantine decisions admit a simple optimal form. If individual $n$ is quarantined, a penalty is incurred only when the individual is not infected, yielding an expected cost of $\alpha_2 (1 - q_n^{(t)})$. If quarantine is not applied, the expected cost corresponds to failing to isolate an infected individual and is equal to $q_n^{(t)}$. Comparing these two quantities shows that quarantine is preferred whenever
\[
q_n^{(t)} > \frac{\alpha_2}{1 + \alpha_2}.
\]

As a result, quarantine decisions follow a fixed threshold rule based solely on the current infection belief and the quarantine cost parameter $\alpha_2$. Importantly, this decision does not depend on how tests are allocated. Quarantine and testing decisions can therefore be decoupled: all methods considered in this work share the same optimal quarantine rule and differ only in how limited testing resources are coordinated under global budget constraints.

\paragraph{Generalized DQN Observations}

The generalized DQN operates at the individual level and estimates action values for two actions, \texttt{test} and \texttt{no-test}, given belief-based infection predictions and recent epidemiological context. Each individual is represented by a fixed 16-dimensional observation vector constructed from the outputs of the supervised learning encoder and the agent’s recent interaction history.

The first six dimensions encode predicted infection probabilities. Specifically, the observation includes predictions from the previous three days as well as predictions for the next three days, providing both retrospective and short-horizon prospective risk information. This structure allows the DQN to reason about temporal trends in infection likelihood rather than relying on a single snapshot.

The next three dimensions record symptom observations over the past three days, capturing recent clinical information that may not yet be reflected in confirmed test outcomes. These are followed by three dimensions indicating whether the individual was tested during each of the past three days, and another three dimensions indicating the corresponding test results when available. Together, these historical features allow the DQN to account for recent information acquisition and avoid redundant or uninformative testing decisions.

The final dimension of the observation encodes the currently active test cost $\alpha^{\text{active}}_{3,t}$, which is supplied by the global coordination mechanism. By conditioning the local Q-function directly on the test cost, a single generalized DQN can adapt its testing behavior smoothly across different scarcity regimes without retraining.

During training, the quarantine cost parameter is fixed to $\alpha_2 = 0.1$, while the test cost $\alpha_3$ is randomly sampled from the range $[0.0, 0.1]$ at episode resets. This encourages the learned Q-function to generalize across testing cost regimes and supports downstream coordination via a shared cost signal.

\subsection{Global PPO Observations}
\label{app:global_obs}

At each timestep, the PPO controller observes a fixed-dimensional state that combines global system-level information with per-cluster summaries. This representation is designed to capture resource demand, temporal context, and heterogeneity across clusters while remaining scalable as clusters dynamically activate and deactivate.

The global component of the observation has dimension $\texttt{GLOBAL\_DIM}=8$ and summarizes the overall state of the system. It includes the current time normalized by the episode horizon, the fraction of clusters that are currently active, and the total number of active individuals normalized by the maximum possible population. To reflect resource availability, the controller observes the current testing budget normalized by its nominal value, as well as the effective budget per active individual. In addition, the observation includes indicators of resource tightness derived from the previous timestep, such as the ratio between unconstrained testing demand and the available budget, the cost multiplier applied most recently relative to the true test cost, and a binary flag indicating whether the system experienced a budget shortage. Together, these features allow the controller to reason about when testing demand is likely to exceed capacity and whether intervention is necessary.

Each cluster is represented by a feature vector of dimension $\texttt{CLUSTER\_DIM}=17$. For an active cluster, this vector encodes its size (normalized by the maximum cluster size), its age relative to the episode length, and short-term histories of testing activity, symptom prevalence, and positive test outcomes over the previous three timesteps. These quantities are normalized by cluster size to ensure comparability across clusters of different scales, and missing history is explicitly encoded to distinguish newly activated clusters. The observation also includes a cost-related feature proportional to the currently applied test penalty and recent testing volume, which reflects the marginal cost faced by that cluster under the active price signal.

To summarize epidemiological risk, we extract aggregate statistics from the local belief-based observation used by the DQN. Specifically, we include the mean and maximum predicted infection probabilities over individuals for both recent and near-future prediction windows. These statistics provide a compact representation of within-cluster risk without exposing individual-level states. Finally, an explicit activity indicator is appended so that inactive or completed clusters can be represented by zero-valued feature vectors while preserving a fixed input dimension.

Inactive clusters are padded with zero vectors, allowing the PPO controller to process up to $\texttt{n\_max}$ clusters using a shared architecture and masking. This observation design enables the controller to regulate testing usage based on global cost signals and coarse-grained cluster heterogeneity, without requiring access to fine-grained individual-level information.

\subsection{Q-Ranking Policy}
\label{app:qranking}

This appendix provides the pseudocode for the global Q-ranking policy used to enforce the hard testing budget. The procedure is deterministic and does not involve learning. At each timestep, it aggregates marginal testing values computed by local policies, ranks candidate tests globally, and executes testing actions up to the available budget. Importantly, testing is assigned only to individuals with positive marginal value, and the budget is not required to be fully utilized if fewer beneficial testing actions are available.
\begin{algorithm}[tb]
\caption{Global Q-Ranking Policy}
\label{alg:qranking}
\textbf{Input}: Active clusters $\mathcal{A}_t$, local observations $\{o_{n,i,t}\}$, global budget $B$\\
\textbf{Output}: Executed actions $\{a_{n,i,t}\}$
\begin{algorithmic}[1]
    \STATE $\mathcal{C} \gets \emptyset$
    \FORALL{$n \in \mathcal{A}_t$}
        \FORALL{individual $i$ in cluster $n$}
            \STATE $\Delta Q_{n,i} \gets Q_{\phi}(o_{n,i,t}, \texttt{test}) - Q_{\phi}(o_{n,i,t}, \texttt{no-test})$
            \STATE append $(n,i,\Delta Q_{n,i})$ to $\mathcal{C}$
        \ENDFOR
    \ENDFOR
    \STATE $\mathcal{C}^+ \gets \{(n,i,\Delta Q_{n,i}) \in \mathcal{C} \mid \Delta Q_{n,i} > 0\}$
    \STATE sort $\mathcal{C}^+$ by $\Delta Q$ in descending order
    \STATE set all $a_{n,i,t} \gets \texttt{no-test}$
    \STATE $K \gets \min(B, |\mathcal{C}^+|)$
    \FOR{$k = 1$ to $K$}
        \STATE let $(n_k, i_k, \cdot)$ be the $k$-th element of $\mathcal{C}^+$
        \STATE $a_{n_k,i_k,t} \gets \texttt{test}$
    \ENDFOR
    \STATE \textbf{return} $\{a_{n,i,t}\}$
\end{algorithmic}
\end{algorithm}

\section{Experimental Details}
\label{app:exp_details}
 
\subsection{Outbreak Simulator and Parameters}
\label{app:simulator}

\begin{table*}[ht!]
\centering
\resizebox{0.8\textwidth}{!}{
\begin{tabular}{ll}
\toprule
\textbf{Parameter} & \textbf{Value}\\ 
\midrule
Incubation period & Lognormal (mean=1.57 days, std=0.65 days)\\
Infectious period & 7 days (from 2 days before to 5 days after symptom onset)\\
Baseline transmission probability & 0.03 per contact\\
Probability symptomatic given infected & 0.8\\
Probability symptomatic without infection & 0.01 per day\\
Probability asymptomatic infection & 0.2\\
Probability highly transmissive index case & 0.109\\
Infectiousness multiplier (highly transmissive case) & 24.4\\
Test sensitivity (True positive rate) & 0.71\\
Test specificity (True negative rate) & 0.99\\
Contact tracing delay & 3 days\\
Test result reporting delay & 1 day\\
Cluster size distribution & Uniform integer distribution [2, 40]\\
\bottomrule
\end{tabular}}
\caption{Key epidemiological parameters used in the agent-based SARS-CoV-2 simulation. Unless otherwise stated, all parameters follow the configuration reported in Peng 2023}
\vspace{-0.8em}
\label{tab:simulation_details}
\end{table*}

We evaluate all methods using a hierarchical multi-cluster outbreak control simulation environment that extends the agent-based SARS-CoV-2 model introduced by Peng~\shortcite{peng2023using}. The simulator models multiple clusters in parallel, where each cluster corresponds to the second-generation contacts of a confirmed index case and evolves independently over time.

At the cluster level, each cluster is simulated using a local environment that captures individual-level infection dynamics, symptom progression, diagnostic testing, and quarantine. Cluster sizes are randomly sampled from the range $[2,40]$, and each episode spans 30 days, with intervention decisions beginning on day 3. Each individual is represented by a fixed-dimensional observation vector encoding epidemiological belief features, symptom history, testing history, and test outcomes. Quarantine decisions follow a threshold policy adopted from Peng~\shortcite{peng2023using}, while testing decisions are controlled by the learned policy.

The local reward follows the single-cluster objective described in Section~\ref{sec: problem}, penalizing infection spread, unnecessary quarantine, and testing costs. Specifically, the reward is defined as $(-S_1 - \alpha_2 S_2 - \alpha_3 S_3)/N$, where $S_1$ penalizes infectious days prior to quarantine, $S_2$ penalizes unnecessary quarantine, and $S_3$ penalizes the number of tests administered. 

At the system level, a global environment manages multiple clusters simultaneously and enforces a hard global budget constraint on daily testing resources. At each timestep, the global environment observes aggregated system-level features, including the current simulation time, the number and scale of active clusters, and recent testing utilization relative to the global budget. It also receives compact cluster-level summaries such as cluster size, time since activation, recent symptom prevalence, and testing activity. Based on these observations, the global controller outputs a cost multiplier that modulates the perceived cost of testing supplied uniformly to all local policies.

All epidemiological parameters used in the simulator are grounded in empirical estimates reported in the epidemiological literature and public health studies on SARS-CoV-2. Although the environment is simulation-based, its parameterization is designed to reflect real-world transmission dynamics, symptom progression, and testing characteristics rather than synthetic or toy settings. For completeness and reproducibility, we summarize the key simulation parameters in Table~\ref{tab:simulation_details} and refer readers to Table~1 of Peng~\shortcite{peng2023using} for additional details and justifications.

\subsection{Training Details}

We train a hierarchical reinforcement learning system consisting of a generalized Transformer-based Deep Q-Network (DQN) for local decision-making and a Proximal Policy Optimization (PPO) policy for global coordination.
Both models are implemented in JAX and Flax to support efficient parallel computation and scalability in the multi-cluster setting.

\paragraph{Local DQN.}
The local testing policy is implemented using a Transformer-based DQN with three encoder layers. Each layer includes a multi-head self-attention block with four attention heads followed by a feed-forward network, using an embedding dimension of 256. The DQN training pipeline is adapted from the CleanRL framework~\cite{huang2022cleanrl}.

The DQN is trained using off-policy reinforcement learning with a replay buffer of size $2\times10^5$ and a batch size of 512. We use the Adam optimizer with gradient clipping at a maximum norm of 1.0 to stabilize updates. A cosine learning-rate schedule with linear warmup over the first $10^4$ steps is applied, starting from an initial learning rate of $5\times10^{-5}$. Training is performed for up to $5\times10^6$ environment steps.

To support joint adaptation with the supervised learning (SL) encoder, DQN training follows a two-timescale alternating schedule. Specifically, we alternate between phases in which the SL encoder is frozen and the DQN is trained for $3\times10^5$ steps, and phases in which the DQN is frozen while new trajectories are collected to fine-tune the SL encoder. At the end of each SL phase, the encoder is updated for a small number of supervised training epochs and then reused for subsequent DQN training. This alternating schedule allows the DQN to adapt to evolving belief estimates while maintaining stable reinforcement learning updates.

Exploration follows an $\epsilon$-greedy strategy with $\epsilon$ linearly annealed from 1.0 to 0.1 over the first 30\% of training. Unless otherwise stated, all experiments are run with five random seeds.

\paragraph{Global PPO.}

The global controller is trained using Proximal Policy Optimization (PPO)~\cite{schulman2017proximal} to regulate system-wide testing scarcity through a continuous cost multiplier. The policy is parameterized by an actor–critic network with a shared Transformer encoder~\cite{vaswani2017attention} that processes both global system features and per-cluster summaries. Global features and cluster-level features are embedded into a 256-dimensional representation and passed through a two-layer Transformer encoder with eight attention heads. The network supports up to 40 concurrently active clusters, with inactive cluster slots masked during both training and execution.

The actor outputs the mean of a scalar Gaussian distribution over a raw action, which is transformed via a sigmoid function and linearly rescaled to produce a cost multiplier in the range $[m_{\min}, m_{\max}]$. The critic outputs a scalar value estimate for the current global state.

PPO training is performed for $3\times10^6$ environment steps using four parallel environments. Each update collects rollouts of 256 steps per environment and performs four optimization epochs with a minibatch size of 256. We use Generalized Advantage Estimation with discount factor $\gamma=0.99$ and $\lambda=0.90$, and normalize advantages using a running mean and variance estimator. Optimization uses the AdamW optimizer with a learning rate of $3\times10^{-5}$, weight decay $10^{-4}$, and gradient clipping with a maximum norm of 0.5. The PPO clipped objective is used with a clip coefficient of 0.10, value loss coefficient of 0.50, and entropy coefficient of 0.001, together with KL-based early stopping (target threshold 0.15) to improve training stability.

The PPO controller does not directly select individual testing actions. Instead, it adjusts the global cost multiplier supplied to local policies, while final test execution is enforced by the global Q-ranking policy, guaranteeing strict adherence to the hard testing budget.

\paragraph{Compute resources.}
All experiments are conducted on the Ascend Cluster at the Ohio Supercomputer Center (OSC)~\cite{OhioSupercomputerCenter1987}, using NVIDIA A100 GPUs and AMD EPYC processors.

\section{Detailed Comparison Policies}
\label{app: policies}

This appendix provides detailed definitions of all comparison policies used in the experiments.
All policies are evaluated in the same environment and strictly enforce feasibility at every timestep.
Unless otherwise stated, all methods share the same epidemiological dynamics, observation space, and reward definition.

Across all methods, \emph{testing and quarantine decisions are decoupled}.
Quarantine decisions are not learned by any method.
Specifically, all baselines, including learned policies, apply the same threshold-based quarantine policy adopted from Peng~\shortcite{peng2023using}, which isolates individuals whose estimated infection risk exceeds a predefined threshold.
The only exception is \textbf{Symp-AvgRand}, which applies symptom-based quarantine as a representative public health heuristic.
As a result, all methods differ \emph{only} in how limited testing resources are allocated under the global budget constraint, ensuring a fair comparison.

\subsection{Heuristic Baselines}

\paragraph{Symp-AvgRand.}
The global testing budget is evenly divided across all active clusters. Within each cluster, individuals are selected uniformly at random for testing. Quarantine decisions are symptom-based and isolate individuals exhibiting observed symptoms or receiving positive test results. This baseline reflects a commonly used public health heuristic that prioritizes symptomatic individuals without explicit risk modeling.

\paragraph{Thres-AvgRand.}
The global testing budget is evenly allocated across clusters.
Within each cluster, individuals are selected uniformly at random for testing. Quarantine follows the threshold-based rule, allowing us to isolate the effect of threshold-based quarantine from testing allocation.

\paragraph{Thres-SizeRand.}
Testing resources are allocated proportionally to cluster sizes.
Within each cluster, individuals are selected uniformly at random for testing. Quarantine follows the same threshold-based rule. This baseline accounts for heterogeneity in cluster size but does not use any learned value estimates.

\subsection{Value-Based Q-ranking Baselines}

\paragraph{Fixed-$M$-QR.}
A fixed global cost multiplier $M$ is applied to the pretrained generalized DQN, inducing a constant perceived test cost across all clusters. Under this fixed cost signal, the local DQN produces per-individual testing values. Candidate tests from all active clusters are aggregated and selected via the global Q-ranking policy subject to the hard budget constraint. This baseline corresponds to a fixed-price relaxation of the constrained allocation problem.

\paragraph{Bin-$M$-QR.}
This baseline adaptively adjusts the testing cost multiplier using binary search to satisfy the global budget constraint.  At each time step, the algorithm uses binary search to find a multiplier $M$ such that the total number of tests proposed by the local deep Q network does not exceed the test budget. Given the resulting multiplier, final test execution is enforced through global Q-ranking.

Conceptually, Bin-$M$-QR implements a simple price-based coordination mechanism that is closely related to Whittle index ideas \cite{whittle1988restless}. Instead of explicitly computing an index for each individual or cluster, the method adjusts a single global cost multiplier until the total number of tests proposed by the local DQNs matches the available budget.

Explicit Whittle index computation is not practical in our setting. Local states are high-dimensional belief representations produced by neural networks, and the environment dynamics are model-free, making analytical or dynamic-programming–based index derivation infeasible. Bin-$M$-QR avoids these difficulties by relying on the fact that the generalized DQN responds monotonically to changes in the testing cost. By increasing or decreasing the cost multiplier, the method effectively shifts which tests are considered worthwhile, resulting in a global cutoff that balances demand and capacity.

As a result, Bin-$M$-QR provides a strong, non-learning baseline for global coordination. It captures the core intuition of index-based allocation, prioritizing actions by marginal value.

\paragraph{Hier-PPO.}
Our proposed approach uses a PPO-based global controller to dynamically output a testing cost multiplier conditioned on the global system state. Local testing values are computed using the pretrained generalized DQN under the induced perceived cost, and final test execution is performed via the same global Q-ranking policy. Unlike Bin-$M$-QR, which adjusts the multiplier greedily to satisfy the budget, the PPO controller learns to anticipate future resource scarcity and cluster dynamics, enabling adaptive coordination across time.

All learning-based baselines share the same pretrained generalized local DQN and differ only in how the testing cost multiplier is selected.

\section{Additional Experimental Results}
\label{app: fullresults}

\subsection{Performance under Fixed Cost Multipliers}
\label{app: fixed_m_ablation}

\begin{table}[htbp]
  \centering
  \footnotesize
  \setlength{\tabcolsep}{1pt}
  \renewcommand{\arraystretch}{1.1}
  \begin{tabularx}{\columnwidth}{l c *{3}{>{\centering\arraybackslash}X}}
    \toprule
    \multicolumn{5}{c}{Asynchronous activation} \\
    \midrule
    $(\#C,\#B)$ & $m=0.5$ & $m=1.0$ & $m=1.5$ & $m=2.0$ \\
    \midrule
    $(20,40)$  & $-0.62{\pm}0.01$ & $-0.65{\pm}0.02$ & $-0.69{\pm}0.01$ & $-0.70{\pm}0.01$ \\
    $(20,400)$ & $-0.64{\pm}0.01$ & $-0.68{\pm}0.01$ & $-0.71{\pm}0.01$ & $-0.73{\pm}0.01$ \\
    \midrule
    \multicolumn{5}{c}{Synchronous activation} \\
    \midrule
    $(\#C,\#B)$ & $m=0.5$ & $m=1.0$ & $m=1.5$ & $m=2.0$ \\
    \midrule
    $(20,40)$  & $-0.33{\pm}0.01$ & $-0.37{\pm}0.01$ & $-0.35{\pm}0.01$ & $-0.38{\pm}0.01$  \\
    $(20,400)$ & $-0.30{\pm}0.01$ & $-0.35{\pm}0.01$ &$-0.39{\pm}0.01$ & $-0.40{\pm}0.01$  \\
    \bottomrule
  \end{tabularx}
  \vspace{-0.5em}
  \caption{Ablation of fixed-$m$ Q-ranking under asynchronous and synchronous activation.}
  \vspace{-0.5em}
  \label{tab:fixed_m_performance}
\end{table}

Table~\ref{tab:fixed_m_performance} reports an ablation study of fixed-$m$ Q-ranking policies. Here, the global cost multiplier $m$ is held constant throughout each episode and used to regulate testing decisions via global Q-ranking.

Across both asynchronous and synchronous settings, performance degrades monotonically as $m$ increases. When $m$ is small, testing is relatively inexpensive, and the policy performs aggressive testing, leading to better outbreak control. As $m$ grows, the effective test penalty increases, causing the policy to suppress testing and miss informative opportunities to identify infected
individuals, which results in poorer overall outcomes.

Importantly, no single fixed value of $m$ performs well across all settings. Smaller $m$ values risk over-testing, while larger $m$ values lead to systematic under-testing. This sensitivity highlights a fundamental limitation of fixed-cost Q-ranking: choosing an appropriate $m$ requires prior knowledge of the operating regime and cannot adapt to changes in resource demand or outbreak dynamics.

\subsection{Ablation on Cost-Sensitivity Regularization}

\begin{figure*}[t]
\centering
\vspace{-1.5em}

\begin{minipage}{0.45\textwidth}
  \centering
  \includegraphics[width=\linewidth]{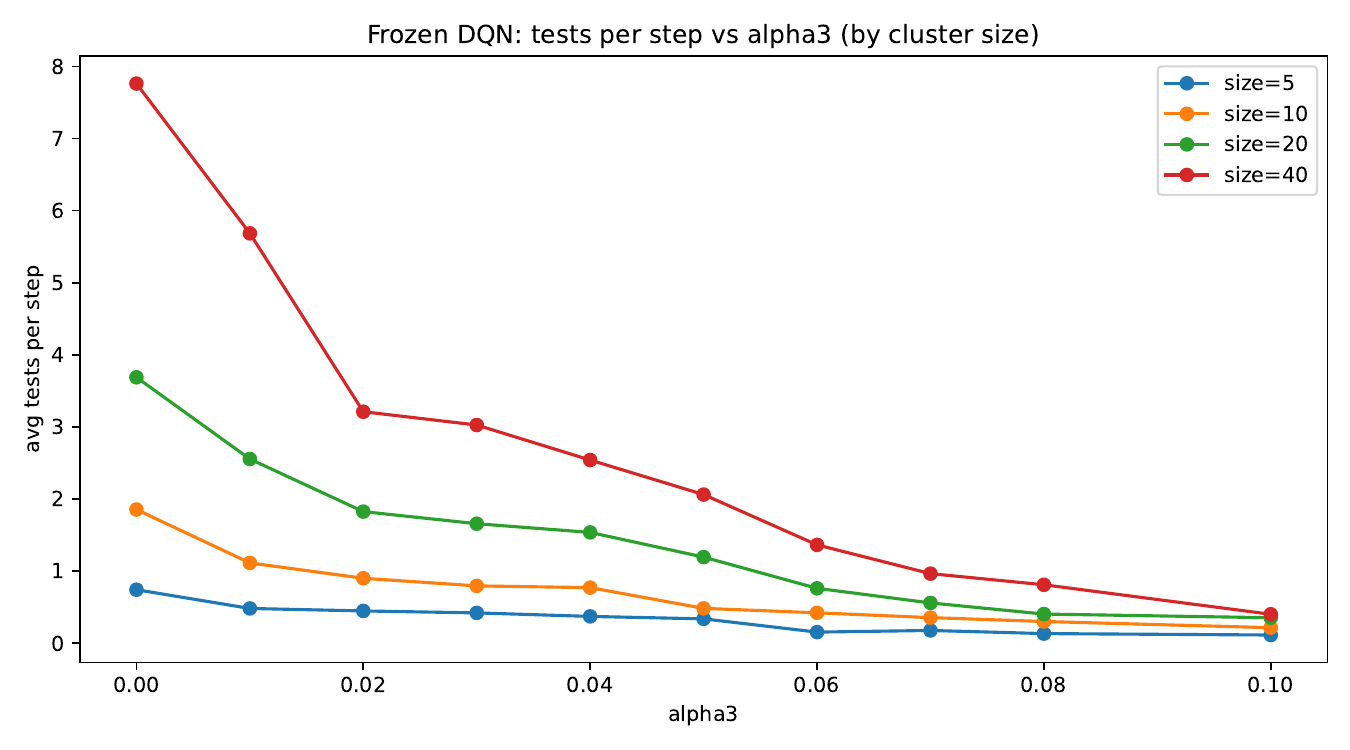}\\
  \small (a) With gradient penalty
\end{minipage}
\hfill
\begin{minipage}{0.45\textwidth}
  \centering
  \includegraphics[width=\linewidth]{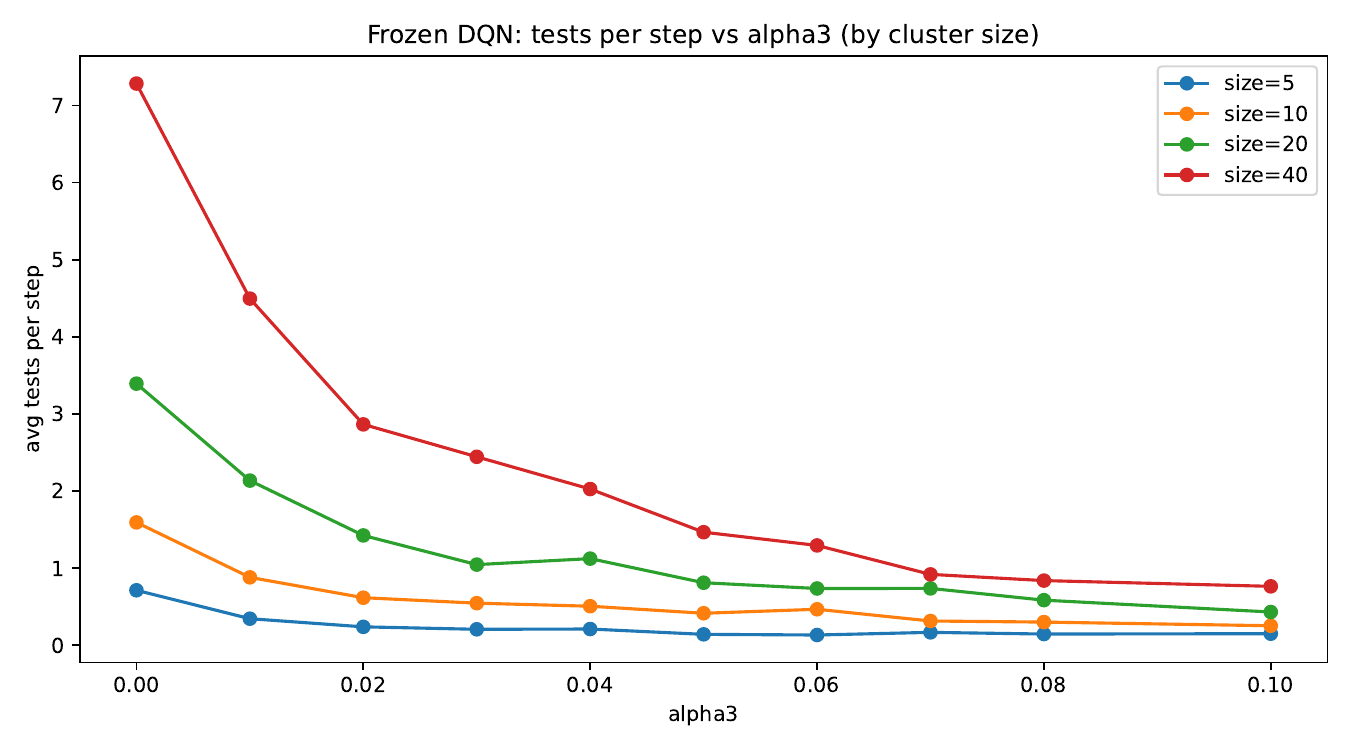}\\
  \small (b) Without gradient penalty
\end{minipage}

\vspace{-0.8em}
\caption{Sensitivity of testing behavior to the test-cost coefficient $\alpha_3$. We report the average number of tests per timestep across different cluster sizes. The gradient-penalty variant exhibits more stable and coherent monotonic responses to increasing $\alpha_3$ than the variant trained without the penalty.}
\label{fig:tests_vs_alpha3}
\vspace{-1em}
\end{figure*}

From Figure ~\ref{fig:tests_vs_alpha3}, we can see that, as expected, increasing the test cost coefficient $\alpha_3$ usually leads to a reduction in the number of tests conducted across all cluster sizes. Therefore, the higher the cost of the test, the more likely the local generalized policy is to avoid testing. However, in the variant trained without gradient penalty, the curve corresponding to cluster size = 10 (green line) exhibits a noticeable non-monotonic outlier. When $\alpha_3$ = 0.04, the average number of tests temporarily increases. This behavior contradicts our explanation for $\alpha_3$, indicating that some unstable situations may occur during the training of the local generalized DQN model.

By contrast, the model trained with gradient penalty shows a stable and consistent monotonically decreasing trend in the test process as $\alpha_3$ increases. The gradient penalty explicitly regulates the sensitivity of Q values relative to $\alpha_3$, thereby enforcing the expected monotonic relationship between test cost and test intensity. As a result, the outlier behavior observed without the penalty is eliminated, and the learned policy exhibits more coherent and stable responses to changes in testing cost.

\subsection{Synchronous vs.\ Asynchronous Cluster Activation}

We report full results under both synchronous and asynchronous cluster activation settings (shown in Table~\ref{tab:full_results}
For cluster counts $\#C \in \{10, 20, 40\}$, we evaluate testing budgets scaled proportionally as $\#B \in \{\,\#C,\,2\#C,\,5\#C,\,20\#C\,\}$.

\subsection{Computational Cost and Runtime}
\label{app:runtime}

Table ~\ref{tab:runtime} reports the average decision time for different numbers of clusters and different test budgets. To eliminate the influence of system size, we fixed the number of individuals in each cluster to 20. As expected, the runtime of both methods increases with the number of clusters. However, Hier-PPO consistently achieves lower decision latency among all the settings, with speedups ranging from approximately $4-8\times$.

The difference in runtime between the two methods is most significant when budgets are tight. When the test budget is relatively small compared to the test demands, Bin-M-QR must repeatedly calculate the multiplier by binary search to enforce the budget constraint, which leads to increased computational overhead. In contrast, Hier-PPO generates the cost multiplier with a single forward propagation of the global policy, thus achieving stable and budget-insensitive runtime.

In summary, these results indicate that while both methods show an approximately linear relationship with the number of clusters. The proposed hierarchical PPO framework is significantly able to make decisions faster, especially in a budget-tight environment.

\begin{table}[t]
\centering
\small
\setlength{\tabcolsep}{1pt}
  \renewcommand{\arraystretch}{1.1}
  \begin{tabularx}{\columnwidth}{l c *{3}{>{\centering\arraybackslash}X}}
    \toprule
    \#C & \#B & Bin-M-QR (ms) & Hier-PPO (ms) & Speedup \\
    \midrule
    10 & 20 & $7.18{\pm}0.47$ & $1.03{\pm}0.12$&  $6.97\times$\\
    10 & 100 & $3.16{\pm}0.56$& $0.61{\pm}0.08$&  $5.18\times$\\
    20 & 40 & $12.39{\pm}0.52$& $2.55{\pm}0.11$&  $4.86\times$\\
    20 & 200 & $4.44{\pm}0.48$& $0.78{\pm}0.15$&  $5.69\times$\\
    40 & 80 & $16.21{\pm}0.59$& $2.64{\pm}0.32$&  $6.14\times$\\
    40 & 400 & $5.87{\pm}0.64$& $0.82{\pm}0.23$&  $7.16\times$\\
    \bottomrule
    \end{tabularx}
\caption{Decision-time runtime comparison under different numbers of clusters. To isolate the effect of cluster count on computational cost, we fix the cluster size to 20 and scale the global testing budget linearly with the number of clusters. Reported values correspond to the average wall-clock decision time per timestep, excluding environment simulation, and include all policy evaluations required for feasibility enforcement.}
\label{tab:runtime}
\end{table}

\subsection{Generalization Across Testing and Quarantine Costs}
\label{app:gen_both}
We further extend the generalized Transformer-based DQN to jointly generalize across both testing and quarantine cost coefficients. In this setting, the observation for cluster $n$ at time $t$ is:
\[
o_{n,i,t} = [\, s_{n,i,t},\;\alpha^{\text{active}}_{2,t},\; \alpha^{\text{active}}_{3,t}],
\]
where $s_{n,i,t}$ is also the individual-level features, including predicted infection probabilities, recent symptom observations, and testing history. By explicitly including both cost coefficients in the observation, a single DQN policy can adapt its testing behavior to different trade-offs between quarantine and testing costs without retraining.

To evaluate the effectiveness of this joint generalization, we compare the generalized DQN against a set of fixed-cost DQN baselines.  For the baselines, each DQN is trained under a single fixed pair of cost coefficients ($\alpha_2$, $\alpha_3$), where $\alpha_2$ is randomly sampled from $[0.0, 0.5]$ and $\alpha_3$ is fixed during training.

Our results show in Table~\ref{tab:gen_both_vs_fixed_dqn}that the jointly generalized DQN achieves performance comparable to, and in some cases indistinguishable from, the corresponding fixed-cost DQNs under matched cost settings. This indicates that conditioning on both $\alpha_2$ and $\alpha_3$ does not introduce noticeable performance degradation, while substantially improving flexibility. 

\begin{table}[htbp]
  \centering
  \small
  \setlength{\tabcolsep}{1pt}      
  \renewcommand{\arraystretch}{1}
  \begin{tabularx}{\columnwidth}{l *{4}{>{\centering\arraybackslash}X}}
    \toprule
        & $\alpha_2{=}0.01$ & $\alpha_2{=}0.1$ & $\alpha_2{=}0.3$ & $\alpha_2{=}0.5$\\
    \midrule
    Fixed-DQN & $-0.39{\pm}0.01$ & $-0.57{\pm}0.02$ & $-0.67{\pm}0.02$ & $-0.83{\pm}0.03$ \\
    Gen-DQN & $-0.41{\pm}0.03$ & $-0.57{\pm}0.03$ & $-0.66{\pm}0.04$ & $-0.86{\pm}0.06$ \\
    \bottomrule
  \end{tabularx}
   \caption{Generalized vs. Fixed DQN performance under varying quarantine cost ($\alpha_2$), with fixed test cost ($\alpha_3$ = 0.05). Results indicate that the generalized policy achieves comparable performance.}
   \label{tab:gen_both_vs_fixed_dqn}
\end{table}

\begin{table*}[t]
\centering

\begin{subtable}{\textwidth}
\centering
\begin{adjustbox}{width=\textwidth}
\begin{tabular}{lcccccc}
\toprule
& Symp-AvgRand & Thres-AvgRand & Thres-SizeRand & Fixed-$M$-QR(1) &Bin-$M$-QR   & Hier-PPO \\
\midrule
\midrule

\#C=10, \#B=10 & $-1.95{\pm}0.05$ & $-0.92{\pm}0.02$ & $-0.98{\pm}0.02$ & $-0.64{\pm}0.02$ & $-0.61{\pm}0.01$ & $\mathbf{-0.56{\pm}0.01}$ \\
\#C=10, \#B=20 & $-2.12{\pm}0.03$ & $-1.02{\pm}0.01$ & $-0.93{\pm}0.02$ & $-0.64{\pm}0.01$ & $-0.62{\pm}0.01$ & $\mathbf{-0.57{\pm}0.01}$ \\
\#C=10, \#B=50 & $-2.64{\pm}0.03$ & $-2.04{\pm}0.01$ & $-2.12{\pm}0.01$ & $-0.64{\pm}0.01$ & $-0.63{\pm}0.02$ & $\mathbf{-0.61{\pm}0.01}$  \\
\#C=10, \#B=200 & $-2.64{\pm}0.03$ & $-2.04{\pm}0.01$ & $-2.12{\pm}0.01$ & $-0.67{\pm}0.01$ & $-0.66{\pm}0.02$ & $\mathbf{-0.63{\pm}0.01}$  \\

\midrule

\#C=20, \#B=20 & $-2.11{\pm}0.04$ & $-0.99{\pm}0.01$ & $-1.03{\pm}0.02$ & $-0.68{\pm}0.01$ & $-0.63{\pm}0.01$ & $\mathbf{-0.58{\pm}0.01}$ \\
\#C=20, \#B=40 & $-2.31{\pm}0.04$ & $-1.11{\pm}0.01$ & $-1.01{\pm}0.02$ & $-0.65{\pm}0.01$ & $-0.64{\pm}0.01$ & $\mathbf{-0.57{\pm}0.01}$ \\
\#C=20, \#B=100 & $-2.47{\pm}0.03$ & $-1.46{\pm}0.01$ & $-1.32{\pm}0.02$ & $-0.68{\pm}0.01$ & $-0.67{\pm}0.01$ & $\mathbf{-0.62{\pm}0.01}$ \\
\#C=20, \#B=400 & $-2.79{\pm}0.02$ & $-2.15{\pm}0.01$ & $-2.21{\pm}0.01$ & $-0.69{\pm}0.01$ & $-0.68{\pm}0.01$ & $\mathbf{-0.63{\pm}0.01}$  \\

\midrule

\#C=40, \#B=40 & $-2.17{\pm}0.03$ & $-1.01{\pm}0.01$ & $-1.04{\pm}0.01$ & $-0.65{\pm}0.01$ & $-0.65{\pm}0.02$ & $\mathbf{-0.58{\pm}0.01}$ \\
\#C=40, \#B=80 & $-2.34{\pm}0.03$ & $-1.13{\pm}0.01$ & $-1.02{\pm}0.01$ & $-0.66{\pm}0.01$ & $-0.67{\pm}0.02$ & $\mathbf{-0.59{\pm}0.01}$ \\
\#C=40, \#B=200 & $-2.53{\pm}0.03$ & $-1.48{\pm}0.01$ & $-1.35{\pm}0.01$ & $-0.69{\pm}0.01$ & $-0.68{\pm}0.02$ & $\mathbf{-0.63{\pm}0.01}$ \\
\#C=40, \#B=800 & $-2.82{\pm}0.01$ & $-2.18{\pm}0.06$ & $-1.56{\pm}0.02$ & $-0.78{\pm}0.01$ & $-0.68{\pm}0.02$ & $\mathbf{-0.64{\pm}0.01}$  \\

\bottomrule
\end{tabular}
\end{adjustbox}
\caption{Asynchronous Activation Environment}
\end{subtable}

\vspace{1em}

\begin{subtable}{\textwidth}
\centering

\begin{adjustbox}{width=\textwidth}
\begin{tabular}{lcccccc}
\toprule
& Symp-AvgRand & Thres-AvgRand & Thres-SizeRand & Fixed-$M$-QR(1) &Bin-$M$-QR  & Hier-PPO \\
\midrule
\midrule
\#C=10, \#B=10 & $-1.08{\pm}0.03$ & $-0.50{\pm}0.01$ & $-0.53{\pm}0.02$ & $-0.38{\pm}0.01$ & $-0.34{\pm}0.01$ & $\mathbf{-0.30{\pm}0.01}$ \\
\#C=10, \#B=20 & $-1.15{\pm}0.03$ & $-0.52{\pm}0.01$ & $-0.50{\pm}0.02$ & $-0.36{\pm}0.01$ & $-0.32{\pm}0.01$ & $\mathbf{-0.27{\pm}0.01}$ \\
\#C=10, \#B=50 & $-1.33{\pm}0.03$ & $-0.65{\pm}0.01$ & $-0.54{\pm}0.01$ & $-0.37{\pm}0.01$ & $-0.38{\pm}0.01$ & $\mathbf{-0.35{\pm}0.01}$  \\
\#C=10, \#B=200 & $-1.55{\pm}0.02$ & $-1.26{\pm}0.01$ & $-1.26{\pm}0.01$ & $-0.38{\pm}0.01$ & $-0.37{\pm}0.02$ & $\mathbf{-0.35{\pm}0.01}$  \\

\midrule

\#C=20, \#B=20 & $-1.09{\pm}0.02$ & $-0.52{\pm}0.01$ & $-0.60{\pm}0.01$ & $-0.37{\pm}0.01$ & $-0.35{\pm}0.01$ & $\mathbf{-0.30{\pm}0.01}$ \\
\#C=20, \#B=40 & $-1.18{\pm}0.02$ & $-0.53{\pm}0.02$ & $-0.53{\pm}0.02$ & $-0.37{\pm}0.01$ & $-0.36{\pm}0.01$ & $\mathbf{-0.31{\pm}0.01}$ \\
\#C=20, \#B=100 & $-1.27{\pm}0.02$ & $-0.64{\pm}0.01$ & $-0.53{\pm}0.02$ & $-0.39{\pm}0.01$ & $-0.37{\pm}0.01$ & $\mathbf{-0.33{\pm}0.01}$ \\
\#C=20, \#B=400 & $-1.57{\pm}0.02$ & $-1.15{\pm}0.02$ & $-1.27{\pm}0.01$ & $-0.38{\pm}0.01$ & $-0.38{\pm}0.01$ & $\mathbf{-0.34{\pm}0.01}$  \\

\midrule

\#C=40, \#B=40 & $-1.14{\pm}0.02$ & $-0.53{\pm}0.01$ & $-0.57{\pm}0.01$ & $-0.32{\pm}0.01$ & $-0.32{\pm}0.02$ & $\mathbf{-0.30{\pm}0.01}$ \\
\#C=40, \#B=80 & $-1.17{\pm}0.01$ & $-0.64{\pm}0.01$ & $-0.52{\pm}0.01$ & $-0.37{\pm}0.01$ & $-0.36{\pm}0.02$ & $\mathbf{-0.31{\pm}0.01}$ \\
\#C=40, \#B=200 & $-1.28{\pm}0.01$ & $-0.64{\pm}0.01$ & $-0.54{\pm}0.01$ & $-0.38{\pm}0.01$ & $-0.37{\pm}0.02$ & $\mathbf{-0.34{\pm}0.01}$ \\
\#C=40, \#B=800 & $-1.57{\pm}0.02$ & $-1.15{\pm}0.06$ & $-1.23{\pm}0.01$ & $-0.40{\pm}0.01$ & $-0.37{\pm}0.02$ & $\mathbf{-0.35{\pm}0.01}$  \\

\bottomrule
\end{tabular}
\end{adjustbox}
\caption{Synchronous Activation Environment}
\end{subtable}
\vspace{-1em}
\caption{Performance comparison of various test allocation methods under synchronous and asynchronous cluster activation environments. Results highlight the advantage of the proposed hierarchical allocation PPO (Hier-PPO) method across different cluster and budget settings. (\#C: number of clusters, \#B: daily testing budget)}
\vspace{-1.5em}
\label{tab:full_results}
\end{table*}

\end{document}